\documentclass{article}

\usepackage{microtype}
\usepackage{graphicx}
\usepackage{booktabs} 
\usepackage{standalone}
\usepackage{tikz}
\usepackage{amsmath}
\usepackage{amsfonts}
\usepackage{physics}
\usepackage{amsthm}
\usepackage{multirow}
\catcode`|=\active
\let|=\mid
\def\yte{y_{\text{query}}}
\def\xte{x_{\text{query}}}
\def\zte{z_{\text{query}}}
\def\ytr{y_{\text{supp}}}
\def\xtr{x_{\text{supp}}}
\def\ztr{z_{\text{supp}}}
\newcommand{\E}{\mathbb E}
\newcommand{\ex}[1]{{\E}_{#1}}
\def\Thetas{\theta,\phi,\psi}
\def\tr{\text{supp}}
\def\te{\text{query}}
\def\f{f_\phi}
\def\q{q_\psi}
\def\p{p_\theta}
\def\NLL{\operatorname{NLL}}

\newif\ifforclass

\usepackage{tabularx}
\usepackage{caption}
\usepackage{subcaption}
\usepackage{float}
\usepackage{hyperref}
\usepackage{cleveref}

\newtheorem{theorem}{Theorem}
\newtheorem{observation}{Observation}

\usepackage[accepted]{icml2020}
\usepackage{bm}

\newcommand{\T}{\mathcal{M}}
\newcommand{\D}{\mathcal{D}}
\renewcommand{\L}{\mathcal{L}}
\newcommand{\N}{\mathcal{N}}
\newcommand{\M}{\mathcal{M}}

\icmltitlerunning{Bayesian Meta-Learning Through Variational Gaussian Processes}

\begin{document}

\ifforclass

\onecolumn
\icmltitle{Bayesian Meta-Learning Through Variational Gaussian Processes}

\icmlkeywords{Machine Learning, Gaussian Processes, Bayesian Meta-Learning}

\vskip 0.3in

\begin{abstract}
\ifforclass

Recent advances in the field of meta-learning have tackled domains consisting of large numbers of small (``few-shot") supervised learning tasks. Meta-learning algorithms must be able to rapidly adapt to any individual few-shot task, fitting to a small support set within a task and using it to predict the labels of the task's query set. This problem setting can be extended to the Bayesian context, wherein rather than predicting a single label for each query data point, a model predicts a distribution of labels capturing its uncertainty. Successful methods in this domain include Bayesian ensembling of MAML-baswed models, Bayesian neural networks, and Gaussian processes with learned deep kernel and mean functions. 

While Gaussian processes have a robust Bayesian interpretation in the meta-learning context, they do not naturally model non-Gaussian predictive posteriors for expressing uncertainty. In this paper, we design a theoretically justified method, VMGP, extending Gaussian-process-based meta-learning to allow for high-quality, arbitrary non-Gaussian uncertainty predictions. Our method uses a latent variable distribution modeled with a Gaussian process along with a learned prediction transformation to model these distributions. Using a learned variational distribution over the latent space, we derive a loss function for meta-training. At meta-test time, the variational distribution allows us to condition on the latent structure of new few-shot tasks, rapidly adapting to the tasks and giving high quality uncertainty estimates.

We compare the predictive posteriors of our method against existing Bayesian meta-learning algorithms on few-shot regression tasks using a negative log-likelihood metric (NLL) that measures both the quality of uncertainty predictions as well as accuracy. On existing benchmark environments with complex non-smooth or discontinuous structure, we find our VMGP method performs significantly better than existing Bayesian meta-learning baselines. We also contribute our own latent function regression environment, and demonstrate that when this environment is composed with complex non-smooth transforms, VMGP's posterior predictions consistently and significantly outperform Bayesian meta-learning baselines.

\begin{figure}[H]
    \centering
    \input{comparison-figures}
    \vskip -2ex
    \parbox\linewidth{
    \begin{tikzpicture}
        \node at (-8, 0) {};
        \node at (8, 0) {};
        \node at (-6.07,0) {EMAML};
        \node at (-1.75,0) {ALPaCA};
        \node at (2.7,0) {DKT};
        \node at (7.05,0) {VMGP (ours)};
    \end{tikzpicture}
    }
  \parbox{\linewidth}{
  \vskip 8pt
  \textit{Figure: Predictions (bottom row) and posterior predictive distribution densities (for the single red testing point, top row) from each model on a $5\operatorname{floor}(z)$ latent function task. VMGP is able to predict a multimodal mixture of point distributions to model its uncertainty, outperforming the other, less expressive methods.}
  }
\end{figure}

\else

Recent advances in the field of meta-learning have tackled domains consisting of large numbers of small (``few-shot") supervised learning tasks. Meta-learning algorithms must be able to rapidly adapt to any individual few-shot task, fitting to a small support set within a task and using it to predict the labels of the task's query set. This problem setting can be extended to the Bayesian context, wherein rather than predicting a single label for each query data point, a model predicts a distribution of labels capturing its uncertainty. Successful methods in this domain include Bayesian ensembling of MAML-based models, Bayesian neural networks, and Gaussian processes with learned deep kernel and mean functions. 
While Gaussian processes have a robust Bayesian interpretation in the meta-learning context, they do not naturally model non-Gaussian predictive posteriors for expressing uncertainty. In this paper, we design a theoretically principled method, VMGP, extending Gaussian-process-based meta-learning to allow for high-quality, arbitrary non-Gaussian uncertainty predictions. 
On benchmark environments with complex non-smooth or discontinuous structure, we find our VMGP method performs significantly better than existing Bayesian meta-learning baselines.

\fi

\end{abstract}
\fi

\twocolumn[

\icmltitle{Bayesian Meta-Learning Through Variational Gaussian Processes}

\begin{icmlauthorlist}
\icmlauthor{Vivek Myers}{stanford}
\icmlauthor{Nikhil Sardana}{stanford}
\end{icmlauthorlist}

\icmlaffiliation{stanford}{Department of Computer Science, Stanford University, Stanford, CA}

\icmlcorrespondingauthor{Vivek Myers}{vmyers@stanford.edu}
\icmlcorrespondingauthor{Nikhil Sardana}{nsardana@stanford.edu}

\icmlkeywords{Machine Learning, Gaussian Processes, Bayesian Meta-Learning}

\vskip 0.3in
] 

\printAffiliationsAndNotice{}

\ifforclass\else
\begin{abstract}

\end{abstract}
\fi

\section{Introduction}

From early childhood, humans have the ability to generalize information across tasks and draw conclusions from few examples. Given a single example of a new object, toddlers can generalize the object's name to others of similar shapes. This ability is not innate, but with only 50--150 objects in their vocabulary, children between 18 and 30 months old learn to master this one-shot classification task \cite{children}. 

Computationally, this problem of generalization is formulated as meta-learning: quickly learning a new task given a set of training tasks which share a common structure. Meta-learning is critical for achieving human-like performance computationally with little data, and recent algorithms have shown success in few-shot image classification and regression problems \cite{maml, reptile}.

However, learning to classify or regress from few examples naturally brings uncertainty. Quantifying and understanding such uncertainty is critical before meta-learning algorithms can be deployed; e.g. autonomous vehicles may be placed in an environment with few prior examples, and must estimate uncertainty to maintain safety and know when to relinquish control. In health care applications, where per-task data is often limited, learning algorithms should estimate uncertainty to gain physicians' trust when patient safety is on the line \cite{healthcare}.

Bayesian methods provide a solution for uncertainty quantification. Rather than output a single label $y$ for an input $x$, Bayesian models assume a prior distribution over their parameters, and predict a posterior distribution $\Pr(y \mid x)$ over the labels which reflects the model's uncertainty in its predictions. Methods for Bayesian supervised learning, such as Bayesian neural networks \cite{bnn} and ensemble models \cite{stein,ensembles} have been extended to meta-learning to provide \textit{task-specific} posterior predictions \cite{bmaml,baysmaml}, building on existing optimization-based meta-learning approaches such as MAML \cite{maml}. 

Gaussian processes (GPs) are a popular Bayesian model that have recently been extended to meta-learning. Gaussian processes allow for a principled way to model covariance between datapoints and produce high-quality uncertainty estimates. Existing Gaussian process-based meta-learning approaches allow for learning the traditionally static, pre-defined kernel and mean function priors by substituting deep networks for them and training across a set of similar tasks \cite{mean,dkt,infotheory,latentgp}.

However, previous GP-based meta-learning approaches do not easily scale to complex non-Gaussian likelihoods. Because these models \cite{mean, dkt} directly use a Gaussian process to predict a probability distribution $\Pr(y \mid x)$ of the labels, their label distribution for any given test input is approximately Gaussian. This can be highly detrimental on regression tasks with discontinuous or less smooth targets.





\subsection{Contributions}
\label{sec:contribution}

In this work, we propose a modification to previous GP-based Bayesian meta-learning approaches, terming our approach a \textbf Variational \textbf Meta-\textbf Gaussian \textbf Process \textbf{(VMGP)}. Rather than fitting a GP directly to the label distribution $\Pr(y|x)$, we use a GP to learn a Gaussian latent variable distribution $\Pr(z|x)$. We then use a deep network to predict labels $y$ from these latent variables $z$. By conditioning the latent variables $z$, at evaluation time, we are able to express and sample from arbitrary non-Gaussian predictive posteriors. We make the following key contributions:
\begin{itemize}
\itemsep 0pt
    \item In \Cref{sec:training}, we derive a new variational loss for training our model, since the addition of the non-Gaussian likelihood mapping the latent space to predictions prevents direct analytical optimization.
    \item In \Cref{sec:evaluation}, we design a principled Bayesian method to condition our latent space on a small support set and generate a predictive posterior distribution over the query labels.
    \item In \Cref{sec:metric}, we describe, motivate, and give a means of approximating the metric, negative log-likelihood (NLL), that we use to compare the uncertainty of Bayesian meta-learning algorithms in regression domains.
    \item In \Cref{sec:functionenv}, we introduce a new function regression dataset environment with more complex functions than standard meta-learning toy regression datasets. We show our model outperforms and achieves more expressive GP-based meta-learning than existing state-of-the-art methods, both on our environment and other standard complex few-shot regression tasks.
\end{itemize}
 

\section{Background}

\subsection{Few-shot Regression}

Formally, in the few-shot learning setting, each task $\mathcal D$ consists of two partitions: the task training set $\D_\tr$ (henceforth referred to as the ``support set") and the task testing test $\D_\te$ (the ``query set"). $\D_\tr = \{(x_j, y_j)\}_{j=1}^{k}$ is a set of $k$ (input, output) pairs, and $\D_\te = \{(x_j, y_j)\}_{j=1}^{m}$ is defined similarly. The tasks are grouped together in a dataset $\T$ and are assumed to be i.i.d samples from the same distribution $p(\T)$. In practice, $k$ is a constant small number (e.g. 5) across all $\D \in \M$, and is referred to as the number of shots.

$\T_{\rm train}$ and $\T_{\rm test}$ are distinct subsets of tasks sampled from $\T$ used for training and evaluating our meta-algorithm, respectively. At meta-train time, the algorithm loops through the tasks $\D \in \T_{\rm train}$, and learns to predict the labels of the task's query set given a query input and the support set $\D_\tr$. During meta-evaluation, we repeat this process for the tasks in $\M_{\rm test}$, except our algorithm does not have access to the ground truth query labels.



\subsection{Bayesian Meta-Learning}
The canonical optimization-based MAML \cite{maml} algorithm operates as follows: We start our model with some pre-trained meta-parameters $\theta$. On each task $\D$, we compute the model loss on $\D_\tr$, take one gradient step w.r.t $\theta$, and compute the loss of the model with these temporary new parameters on $\D_\te$. We sum the losses over all tasks $\D$ and run gradient descent on this sum of losses w.r.t the meta-parameters $\theta$.
\[\min_\theta \sum_{\D \in \T_{\rm train}} \L(\theta - \alpha \nabla_\theta \L(\theta, \D_\tr), \D_\te)\]
At test time, we evaluate the query inputs using the fine-tuned parameters: $\phi \gets \theta - \alpha\nabla_\theta \mathcal{L}(\theta, \D_\tr)$.

We can convert MAML into a Bayesian meta-learning model simply by running an ensemble of them (EMAML). For each testing point $x_{\rm query}$, we treat the $i$-th MAML's predicted label $\hat{y}_{i} = f_{\phi_i}(x_{\rm query})$ as a sample from the ``posterior distribution" of MAML outputs given $x_{\rm query}$. Further enhancments on EMAML include BMAML \cite{bmaml}, which, among other improvements, uses Stein Variational Gradient Descent \cite{stein} to push each of its MAMLs apart to ensure model diversity.

ALPaCA is another Bayesian meta-learning algorithm for regression tasks \cite{alpaca}. ALPaCA can be viewed as Bayesian linear regression with a deep learning kernel. Instead of determining the MAP parameters for $y_i =\theta^\top x_i + \varepsilon_i$, with $\varepsilon_i \sim \N(0, \sigma^2)$, as in standard Bayesian regression, ALPaCA learns Bayesian regression with a basis function $\phi:\mathbb{R}^{|x|} \to \mathbb{R}^{|\phi|}$, implemented as a deep neural network. Thus, $y$ is regressed as $y^\top = \phi(x)^\top\bm{K} + \varepsilon$, for $\varepsilon \sim \N(0, \bm{\Sigma}_\varepsilon)$, $\bm{K} \in \mathbb{R}^{|\phi| \times |y|}$. $\bm{K}$ has a prior of a matrix normal distribution \cite{wu2020alpaca}. As with standard Bayesian linear regression, ALPaCA is able to produce Gaussian predictive posteriors quantifying its uncertainty at evaluation. 

In our results section, we use EMAML and ALPaCA as baselines. We only compare against one MAML-based model because past work has shown MAML-based methods to generally be less competitive on environments similar to the few-shot regression ones we run experiments on \cite{dkt}.

\subsection{Gaussian Processes}

In a Gaussian process, we assume there exists some unknown function $f$ that relates our inputs $x$ and our labels $y$ with noise: $y = f(x) + \varepsilon$. We further assume the distribution $p(f \mid \bm{x})$ over such functions given any finite set $\bm{x}$ of inputs is a multivariate Gaussian with a prior mean and covariance (kernel) function. During training, the Gaussian process analytically produces the distribution $p(f \mid \bm{x}, \bm{y})$ over functions $f$ by conditioning on a training dataset $(\bm{x}, \bm{y})$. At test time, we take our testing data points $\bm{x^*}$, sample a function $\hat{f} \sim p(f \mid \bm{x}, \bm{y})$, and analytically compute the posterior distribution over our predicted labels $\Pr(\bm{\hat{y}^*} \mid \bm{x}, \bm{y}, \bm{x^*})$ \cite{gptextbook}.


\subsection{Bayesian Meta-Learning with Gaussian Processes}
Gaussian processes have a natural meta-learning interpretation---instead of pre-defining a mean and covariance function, we learn them across a set of tasks. Fitting the mean and kernel functions (the prior) corresponds to meta-training, while evaluation is performed by conditioning on the $\mathcal D_\tr$ of a given task. 

Previous GP-based meta-learning works have replaced the mean and covariance functions with deep neural networks. Fortuin \& R\"{a}tsch \yrcite{mean} found improvements on step-function regression tasks using a learned mean function. Patacchiola et al. \yrcite{dkt} reported strong accuracy improvements and quantitatively estimated uncertainty on toy regression, facial pose estimation, and few-shot image classification datasets with their learned deep kernel (DKT) model. 

We use DKT with an RBF kernel as a baseline in our results section.

\subsection{Variational Inference}

The core goal of our method proposed in \Cref{sec:contribution} is to be able to express arbitrary non-Gaussian prediction distributions. We accomplish this goal by learning an unobserved latent variable distribution $z$ that adheres to an analytically tractable multivariate Gaussian distribution. Thus, variational inference, which aims to learn latent variables $z$ given observed variables, is invaluable \cite{variational}. 

In particular, we adapt the common variational approach of maximizing an evidence lower bound (ELBO) on the log likelihood of data with a learned  variational posterior $q$,\begin{equation*}
    \log p(x)\geq \ex{q(z|x)}\log p(x|z) - D_{KL}[q(z|x)\|p(z)].
\end{equation*}
Past work has used variational approaches to learn approximate GPs on large datasets using inducing point methods \cite{svgp,deepgp,gptextbook}. Unlike these methods, we focus on few-shot regression, and thus are able to use exact GPs. While exact GPs usually do not train on enough data to make variational inference viable, in our setting, even though we only condition our GPs on small support sets, during meta-training, our model is exposed to enough training data across tasks to learn sophisticated latent structure.

\section{Approach}

Similar to Patacchiola et al. \yrcite{dkt}, we view the meta-learning process as consisting of two steps:\begin{enumerate}
    \item Meta-training. Our model learns to maximize the predicted likelihood of each task's labels $y$ given the datapoints $x$ (Type-II MLE estimation).
    \item Meta-evaluation. Our model predicts the conditional distribution of the task's labels $\yte$ for the unlabeled 
    query set $\xte$ given the labeled support set $(\xtr, \ytr)$.
\end{enumerate}

Unlike past approaches, our model's predictive posterior $\Pr(y\mid x)$ is not constrained to be Gaussian. Rather, we learn a set of latent variables $z$ modeled by a multivariate Gaussian. Intuitively, all task-specific structure will ``factor" through these latent variables. By requiring them to follow a multivariate Gaussian, we gain the ability to analytically condition their values for a given task, which we will show is essential for prediction. 

From this latent space, we learn a mapping $f(y|z)$ to the predicted labels. Two immediate challenges present themselves: \begin{enumerate}
    \item Directly computing the likelihood $\Pr(y|x)$ for meta-training now requires the intractable integral $\int_z \Pr (y|z,x)\Pr(z|x)\,\dd z$.
    \item Adapting to the support set $(\xtr,\ytr)$ at meta-evaluation requires conditioning on the latent values $\ztr$. However, we have \textit{a priori} no way to get the latent values for the support set from $\xtr$ and $\ytr$.
\end{enumerate}

We solve both of these problems using techniques from variational inference. By maintaining a variational distribution $q(z|x,y)$, we are able to obtain an ELBO training objective, and at evaluation, sample possible latent variables for the support set.

\begin{figure}
    \centering
    \begin{tikzpicture}[every node/.style={scale=1.5},fn/.style={fill=gray!20,rounded corners},input/.style={-latex,black!30!green,thick},output/.style={-latex,red,thick}]
	\node (x) at (0,0) {$x$};
	\node[fn] (p) at (2,-.5) {$p_\theta$};
	\node[yshift=-3ex,text width=80,scale=.45] at (p) {fixed mean deep kernel GP};
	\node (z) at (4,-1) {$z$};
	\node (y) at (8,0) {$y$};
	\draw[input] (x) -- (p);
	\draw[output] (p) -- (z);
	\node[fn] (f) at (6,-.5) {$f_\phi$};
	\node[fn] (q) at (4,2) {$q_\psi$};
	\draw[input] (x) -- (q);
	\draw[output] (q) -- (z);
	\draw[input] (z) -- (f);
	\draw[input] (y) -- (q);
	\draw[output] (f) -- (y);
	\node[yshift=3.5ex,text width=55,scale=.45] at (q) {deep mean /kernel GP};
	\node[yshift=-3.3ex,text width=80,scale=.45,xshift=3.2em] at (f) {MLP};
\end{tikzpicture}
    \caption{Graphical model of the components of our approach. Red arrows are outputs and green arrows are inputs. Each boxed node is a learned distribution over its output conditioned on the input variables.}
    \label{fig:model}
\end{figure}
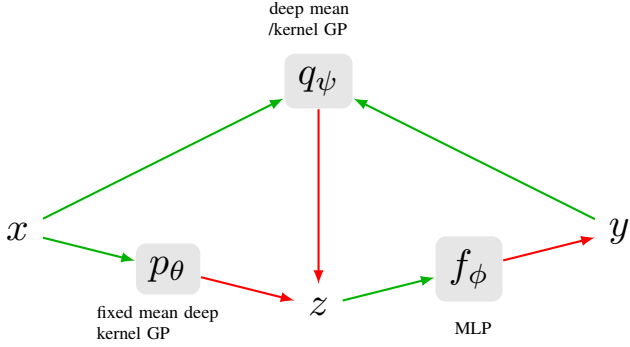
\subsection{Derivation}
\label{sec:formulation}

During meta-training, we are presented with a series of tasks $(x, y)\sim \T$. We assume there is some latent structure to the tasks, $z$, such that for each task $\Pr(z|x)$ is a multivariate Gaussian and $\Pr(y|x,z)=\Pr(y|z)$ is a diagonal Gaussian $\mathcal N(\cdot, \varepsilon I).$

Our goal during meta-training is to learn the distribution $\Pr(y|x)$. As an optimization problem, we want to learn a parameterized model $P_\Theta$ minimizing $D_{KL}[\Pr(y|x)\| P_\Theta(y|x)]$. Indeed, we see that this optimization corresponds to maximizing the quantity $\log P_\Theta(y|x)$. Proceeding, and adding a parameterized multivariate Gaussian latent prior $p_\theta(z|x)$, we see
\begin{align*}
    \log P_\Theta(y|x)&=\log \ex{p_\theta(z|x)} P_\Theta(y|z,x)
\end{align*}
Applying the standard variational ELBO bound with a learned variational multivariate Gaussian distribution $q_\psi(z|x,y)$, and noting from our latent assumption $\Pr(y|x,z)=\Pr(y|z)$ that we can write $f_\phi(y|z):=P_\Theta(y|z,x)$, we get (cf. Kingma \& Welling \yrcite{vae}, Blei et al. \yrcite{variational}, or Hensman et al. \yrcite{svgp}):
\begin{align}
&\log\,\ex{\p(z|x)} P_\Theta(y|z,x) \nonumber \\
&=
\log\,\ex{q_\psi(z|x,y)} {p_\theta(z|x) \over q_\psi(z|x,y)}f_\phi(y|z)\nonumber \\
    &\geq \ex{q_\psi(z|x,y)} \log {p_\theta(z|x) \over q_\psi(z|x,y)}f_\phi(y|z)\nonumber \\
    &=\ex{q_\psi(z|x,y)}\log f_\phi(y|z) - D_{KL} [q_\psi(z|x,y)\|p_\theta(z|x)] \label{eq:elbo}
\end{align}
Now, to learn the model $P_\Theta$ we must simply maximize the bound from \Cref{eq:elbo} with respect to the parameters composing $\Theta$, namely $\Thetas$. See \Cref{fig:model} for a diagram of the learned distributions. 

\subsection{Training}
\label{sec:training}

\begin{algorithm}
   \caption{Meta-Training}
   \label{alg:training}
\begin{algorithmic}[1]
   \STATE {\bfseries Input:} meta-train set $\T_{\rm train}$
   \STATE \hphantom{\bfseries Input:} model $P_\Theta=(\p,\f,\q)$
   \WHILE {training}
   \STATE $\{\D\}_{i=1}^{N_{\rm batch}}\sim \T$
   \STATE $x_i, y_i \gets \D_i$
   \STATE $\Theta\gets \Theta-\alpha\grad_\Theta\mathcal L(x,y;\Thetas)$\hfill{ \cref{eq:miniloss}}
   \ENDWHILE
\end{algorithmic}
\end{algorithm}

During training, we maximize the bound from \Cref{eq:elbo} across all training tasks. As such, we obtain a meta-training loss function for a single task: \begin{align}
    &\mathcal L(x, y; \Thetas) = \nonumber\\&
     D_{KL} [q_\psi(z|x,y)\|p_\theta(z|x)] - \ex{q_\psi(z|x)}\log f_\phi(y|z),
     \label{eq:miniloss}
\end{align}
and thus an overall loss of \begin{align}
    \mathcal L'(\Thetas)=\ex{(x,y)\sim\T_{\rm train}} \,\mathcal L(x, y; \Thetas)
         \label{eq:loss}.
\end{align}
We minimize the loss in \Cref{eq:loss} through gradient descent with respect to $\Thetas$, as shown in \Cref{alg:training}.

Intuitively, the $D_{KL} [q_\psi(z|x,y)\|p_\theta(z|x)]$ term in \Cref{eq:miniloss} encourages $\p$ to be a good approximation of $\q$ that does not use information from $y$, while simultaneously encouraging $\q$ to use less information from $y$ to allow $\p$ to be a good approximation of it. This similarity is essential at evaluation time  (\Cref{sec:evaluation}) for modeling the relationship between the latent structure of support and query sets. We can analytically compute this term using the closed form KL-divergence between multivariate Gaussians:
\begin{align*}
&D_{\mathrm{KL}}\left[\mathcal{N}_{0}(\mu_0,\Sigma_0) \| \mathcal{N}_{1}(\mu_1,\Sigma_1)\right]
=\frac{1}{2}\biggl[\operatorname{tr}\left(\Sigma_{1}^{-1} \Sigma_{0}\right)\\&\qquad+\left(\mu_{1}-\mu_{0}\right)^{\top} \Sigma_{1}^{-1}\left(\mu_{1}-\mu_{0}\right)-k+\ln \left(\frac{\operatorname{det} \Sigma_{1}}{\operatorname{det} \Sigma_{0}}\right)\biggr],\end{align*} with $k=|\!\D\!|$ the dimension of the distributions.

Meanwhile, the $-\ex{q_\psi(z|x)}\log f_\phi(y|z)$ term can be seen as an $\ell_2$ predictive loss (modeling $f_\phi(y|z)$ as a fixed-variance diagonal Gaussian), encouraging the model to accurately predict labels $y$ from the latent space $z$, which has obvious utility for the model's performance.

\begin{algorithm}
   \caption{Meta-Testing}
   \label{alg:testing}
\begin{algorithmic}[1]
   \STATE {\bfseries Input:} meta-test task $(\D_\tr,\D_\te)=\D\sim\T_{\rm test}$
   \STATE \hphantom{\bfseries Input:} model
   $P_\Theta=(\p,\f,\q)$
   \STATE {\bfseries Output:} samples from predictive posterior
   \STATE $x_\tr, y_\tr \gets \D_\tr$
   \STATE $x_\te, \bullet \gets \D_\te$
   \STATE $R \gets \{\}$
   \FOR {$i\in\qty{1\ldots N_{\rm samples}}$}
   \STATE $z_\tr\sim\q(z_\tr|x_\tr,y_\tr)$\hfill \cref{eq:trsamp}
   \STATE $z_\te\sim\p(z_\te|x_\te,z_\tr,x_\tr)$\hfill \cref{eq:zsamp}
   \STATE $y_\te\sim\f(y_\te|z_\te)$\hfill \cref{eq:ysamp}
   \STATE $R\gets R\cup \qty{y_\te}$
   \ENDFOR
   \RETURN $R$
\end{algorithmic}
\end{algorithm}
\subsection{Evaluation}
\label{sec:evaluation}

At evaluation, our model should for a given task support set $(\xtr, \ytr)$ and query datapoints $\xte$, predict the query labels $\yte$. In other words, the desired prediction (``predictive posterior") is $\Pr(\yte|\xte,\ytr,\xtr)$. To avoid requiring a restrictive analytic expression for this distribution, we merely require that our algorithm produce samples $\yte \sim\Pr(\yte|\xte,\ytr,\xtr)$ from the posterior.

To generate a sample $$\yte \sim P_\Theta(\yte|\xte,\ytr,\xtr)$$ where as before $P_\Theta$ is our learned approximation of the true $\Pr(y|x)$, it suffices to generate a sample from the joint distribution $P_\Theta(\yte, \zte, \ztr|\xte,\ytr,\xtr)$. Applying the probability chain rule, we can approach this sampling iteratively, 
\begin{flalign}
    \text{first taking}  && \ztr&\sim q_\psi(\ztr|\xtr,\ytr) \label{eq:trsamp},\\
    \text{then}  && \zte&\sim p_\theta(\zte|\xte,\ztr,\xtr), \label{eq:zsamp}\\
    \text{and finally}  && \yte&\sim f_\phi(\yte|\zte) \label{eq:ysamp}.
\end{flalign}

The final $\yte$ is the desired sample from posterior. Note that the distribution $p_\theta(\zte|\xte,\ztr,\xtr)$ in \Cref{eq:zsamp} is obtained by conditioning the joint distribution $p_\theta(\zte,\ztr|\xte,\xtr)$ on the value of $\ztr$ sampled in \Cref{eq:trsamp}. This conditioning can be done analytically precisely because of our construction of $p_\theta$ as a multivariate Gaussian distribution. Indeed, it is a well-known result of the multivariate Gaussian \cite{gptextbook} that if we can write \begin{equation*}
    p_\theta(\zte,\ztr|\xte,\xtr)=
    \mathcal N\qty(\mqty[\mu_1\\\mu_2],
    \mqty[\Sigma_{11}&\Sigma_{12}\\
    \Sigma_{21}&\Sigma_{22}])
\end{equation*} we can say by conditioning that \begin{align*}
    &p_\theta(\zte|\xte,\ztr,\xtr)\\&=\mathcal N(
    \mu_1+\Sigma_{12}\Sigma_{22}^{-1}(z_\tr - \mu_2),\Sigma_{11}-\Sigma_{12}\Sigma_{22}^{-1}\Sigma_{21}
    ).
\end{align*}
We note the critical role each of the three distributions $\p, \f, \q$ trained using \Cref{eq:loss} in our method. Using $\q$, we are able to extract the latent structure $\ztr$ of the support task $(\xtr,\ytr)$. Then, using the covariance between the latent variables modeled by $p_\theta$, we are able to obtain the latent structure $\zte$ of the query task $\xte$ from the latent structure of the support task $\ztr$. (Notably, we cannot use $q_\psi$ to obtain $\zte$ since $q_\psi$ takes $y$ as an input, which we do not have for the query set.) Finally, with the model $\f$, we map from the sampled latent structure $\zte$ of the query set to a sampled value for $\yte$. This method is summarized in \Cref{alg:testing}.

\subsection{Architecture}
\label{sec:architecture}

To implement our algorithms in \Cref{sec:training} and \Cref{sec:evaluation}, we need differentiable parameterized models for $\p,\f,$ and $\q$. Both $\p$ and $\q$ output multivariate Gaussians, and as such the natural choice is to model them as Gaussian processes. To allow maximal expressiveness, both $\p$ and $\q$ use deep kernel functions consisting of a learned embedding using a multilayer perceptron (MLP) model composed with an RBF kernel, a popular architecture for regression tasks \cite{dkt,mean}. 

Since $\p$ represents a prior over the latent variables (it does not condition on $y$), it uses a constant mean function to avoid overfitting to the biases of the $\xtr$ in the training tasks. However, since $\q$ represents a posterior that is already conditioned on $y$, $\q$ needs the ability express this conditional distribution which likely does not have a constant mean, and so $\q$ uses a deep mean function predicted by an MLP. 

Noting the assumed structure of $\f$ in \Cref{sec:formulation} (so all the covariance between $y$ factors through $z$), $\f$ should produce a diagonal Gaussian with small fixed variance $\varepsilon/2$ over $y$ given values of $z$. So, we can equivalently represent $\f$ as a deterministic MLP mapping $z\to y$, and view the $\log \f(y|z)$ term in \Cref{eq:loss} as an $\ell_2$ loss $\varepsilon^{-1}\|f(z)-y\|_2^2$ (where the hyperparameter value $\varepsilon=0.01$ was found to empirically be effective).

\subsection{Uncertainty Metric}
\label{sec:metric}

At test-time, we validate our model on testing tasks with $\mathcal D_\tr = (\xtr, \ytr)$ and $\mathcal D_\te = (\xte,\yte)$. Using \Cref{eq:trsamp,eq:zsamp,eq:ysamp} we are then able to generate samples from the model's predictive posterior:
\begin{equation}
    \yte\sim P_\Theta(\yte|\xte,\ytr,\xtr).
\end{equation}
To compare the uncertainty represented by the predictive posteriors of different Baysesian meta-learning algorithms, $P(\yte|\xte,\ytr,\xtr)$, we require a standardized metric that can be computed using only samples from the predictive posterior and the true value of $\yte$, which we denote $\yte^*$. A ``good" Bayesian meta-learning algorithm should predict a posterior \[P(\yte) \approx \Pr(\yte|\xte,\ytr,\xtr)\] under which the true value $\yte^*$ has high probability $P(\yte^*)$. 

A standard metric that stratifies our desideratum is the negative log-likelihood metric, used for evaluating many Bayesian learning algorithms \cite{alpaca,million,dspp}. We can define the NLL metric as follows,
\begin{equation}
    \NLL(P,y^*)=-\log P(y^*), \label{eq:nll}
\end{equation}
where $P(y)$ is the predictive posterior of a Bayesian algorithm about some datapoint $x$ with true value $y^*$. 

As noted previously, we cannot assume there is an analytic form to the predicted posteriors of our algorithms, making the PDF term $P(y^*)$ in \Cref{eq:nll} intractable. We propose the following approximation for \Cref{eq:nll}:
\begin{equation}
    \NLL_\xi(P,y^*)=-\log \ex{y\sim P(y)}(\xi\pi)^{-1/2}e^{-\xi^{-1}(y^*-y)^2},\label{eq:nlle}
\end{equation}

which can now be computed through Monte Carlo sampling $y\sim P(y)$. We can view this computation of \Cref{eq:nlle} with samples $y_1\ldots y_N \sim P(y)$ as approximating $P(y)$ as a uniform mixture of the Gaussians $\mathcal N(y_i,\xi/2).$

\begin{theorem}\label{thm:converge} Consider a fixed piecewise continuous probability density $P(y)$. We have
$\NLL_\xi(p, y^*) \to \NLL(P, y^*)$ for $y$ a.e. as $\xi\to0$.
\end{theorem}
\begin{proof}
Take any $y$ in the a.e. set where $P(y)$ is continuous. 
\allowdisplaybreaks
\begin{align*}
    &\NLL_\xi(P,y^*) \\&= -\log \ex{y\sim P(y)}(\xi\pi)^{-1/2}e^{\xi^{-1}(y^*-y)^2}\\
    &= -\log\int_y P(y) (\xi\pi)^{-1/2}e^{\xi^{-1}(y^*-y)^2}\,\dd y.
\end{align*}
We know that $(\xi\pi)^{-1/2}e^{\xi^{-1}(y^*-y)^2}$ is a Gaussian density with variance shrinking as $\xi\to0$. Taking $\xi$ small, an arbitrarily close to $1$ portion of the mass of the Gaussian can be contrained to be within a radius-$\varepsilon$ interval of $y$. So, by continuity of $P$ at $y$, this last expression goes to $-\log P(y^*)$ as $\xi\to0$, and we see $\lim_{\xi\to0}\NLL_\xi(P,y^*)=\NLL(P,y^*).$ 
\end{proof}

\begin{theorem}\label{thm:sum}
For a continuous random variable $X$ with density $P_X$, 
$\NLL_\xi(P_X, y^*)=\NLL(P_{X+\varepsilon}, y^*)$ where $\varepsilon\sim\mathcal N(0, \xi/2).$
\end{theorem}
\begin{proof}
We see \begin{align*}
    &\NLL_\xi(P_X,y^*) \\&= -\log \ex{y\sim P_X(y)}(\xi\pi)^{-1/2}e^{\xi^{-1}(y^*-y)^2}\\
    &= -\log\int_y P_X(y) P_\varepsilon(y^*-y)\,\dd y\\
    &=  -\log(P_X * P_\varepsilon)(y^*) \\
    &= -\log P_{X+\varepsilon}(y^*) =\NLL(P_{X+\varepsilon}, y^*) 
\end{align*}
where $*$ denotes the convolution operation.
\end{proof}

In light of \Cref{thm:sum}, the $\NLL_\xi(P, y^*)$ measures the true $\NLL$ of a predicted posterior $P$ at $y^*$, perturbed by Gaussian noise with variance $\xi/2$. Taking $\xi$ small, the perturbation becomes minimal, and $\NLL_\xi$ becomes the $\NLL$ of a distribution $P'$ that is almost equal to $P$. Indeed, by \Cref{thm:converge}, for $\xi$ sufficiently small, $\NLL_\xi(P, y^*)$ converges to the true $\NLL_\xi(P, y^*)$. Thus, dropping the constant terms in \Cref{eq:nlle}, we select as our final metric\begin{align}
    \widehat{\NLL}(P,y^*)=-\log \ex{y\sim P(y)} e^{-\xi^{-1}(y^*-y)^2},\label{eq:metric}
\end{align} for a small choice of $\xi$, computed with Monte Carlo samples from the posterior. Empirically, we find that since the tasks in our experiments do not exhibit pathological behavior, using $\xi=0.1$ in \Cref{eq:metric} yields good, low-variance $\NLL$ approximations.

\subsection{Analysis} 
We can view the MLP models used in \Cref{sec:architecture} as universal function approximators, able to learn continuous functions arbitrarily well \cite{universal}. Past work on GPs for meta-learning \cite{dkt,mean,alpaca} have used universal function approximators to model mean and covariance functions, but then are only able to produce Gaussian predictive posteriors. Meanwhile, ensemble methods \cite{bmaml} are able to model non-Gaussian posteriors, however, they provide a less direct Bayesian justification for their predictions compared to Gaussian processes.

\begin{observation}
Any distribution $\Pr(y|x)$ with continuous density can be factored through our latent variables $z$ under a deterministic continuous mapping $f(z)=y$ and $\Pr(z|x)$ a multivariate Gaussian. \label{obs:mapping}
\end{observation}
\begin{proof}
Indeed, suppose we have a one dimensional latent space $\Pr(z|x)=\mathcal N(0,1)$. Let $\Phi$ be the CDF of $z$ and and $\Psi$ the CDF of $\Pr(y|x)$. Defining $f(z)=\Psi^{-1}(\Phi(z))$, it is easy to see that for $z\sim\mathcal N(0,1)$, $f(z)$ is distributed according to $\Pr(y|x)$. Further, $f$ is a composition of CDFs of continuous densities, and thus continuous.
\end{proof}
Now, for any true task posterior $\Pr(y|x)$, we know from Observation~\ref{obs:mapping} there exists a continuous map $f$ from a Gaussian latent space that is able to represent $\Pr(y|x)$. In \Cref{sec:architecture}, we learn the map $\f:z\to y$ using an MLP. Since $\f$ can be viewed as a universal function approximator, we then see, for correctly chosen optimization parameters and architecture, $\f$ will be able to learn $f$, and our model will converge to predicting true task posteriors.

Thus, our model extends existing GP-based meta-learning algorithms by theoretically allowing arbitrary non-Gaussian posterior prediction distributions to be learned in a principled way.

\section{Results}

\begin{table*}[htbp]
\begin{center}
    \begin{tabular}{l c c c c}
        \toprule
        & $\arctan(1/z)$ & $5\operatorname{floor}(z)$ & $\tan(z)$ & $5\sin(1/z)$ \\
        \midrule
        EMAML & $5.474 \pm 0.011$ & $63.621 \pm 1.095$ & $85.602 \pm 1.794$ & $71.703 \pm 0.804$ \\
        ALPaCA & $1.979 \pm 0.038$ & $4.559 \pm 0.087$ & $23.677 \pm 0.848$ & $7.994\pm 0.171$\\
        DKT & $1.851 \pm 0.040$ & $5.270 \pm 0.125$ & $19.704 \pm 0.664$ & $7.019 \pm 0.141$ \\
        VMGP (ours) & $\bm{1.593 \pm 0.047}$ & $\bm{3.990 \pm 0.208}$ & $\bm{16.867 \pm 1.086}$ & $\bm{4.784 \pm 0.122}$ \\
        \bottomrule
    \end{tabular}
\end{center}
\caption{Average NLL $\pm$ Std. Error for Latent Gaussian Environment Datasets.}
\label{tab:latentnll}
\end{table*}
\begin{table*}[htbp]
\begin{center}
    \begin{tabular}{l c c c c}
        \toprule
        & Standard & High Frequency & Out of Range & Tangent \\
        \midrule
        EMAML & $24.565 \pm 0.607$ & $37.416 \pm 0.818$ & $43.173 \pm 2.491$ & $219.335 \pm 4.047$\\
        ALPaCA & $\bm{2.300 \pm 0.091}$ & $4.117 \pm 0.081$ & $\bm{0.702 \pm 0.073}$ & $44.021 \pm 1.829$\\
        DKT & $3.167 \pm 0.146$ & $3.817 \pm 0.079$ & $2.785 \pm 0.154$ & $35.997 \pm 1.142$ \\
        VMGP (ours) & $3.366 \pm 0.159$ & $\bm{3.720 \pm 0.142}$ & $3.422 \pm 0.385$ & $\bm{19.500 \pm 2.042}$ \\
        \bottomrule
    \end{tabular}
\end{center}
\caption{Average NLL $\pm$ Std. Error for Sinusoid Environment Datasets.}
\label{tab:sinnll}
\end{table*}

\subsection{Experimental Details}

We compared our VMGP method (Figure \ref{fig:model}) against the three baselines mentioned in Section 2: Deep Kernel Transfer with an RBF Kernel (henceforth referred to as DKT), ALPaCA, and EMAML with 20 one-step MAML particles \cite{dkt, alpaca, bmaml}.

All models were trained with a backbone MLP with 2 hidden layers, each with 40 units followed by a ReLU activation. MAML inner learning rates were set to $0.1$. All models used Adam for optimization with a learning rate of $10^{-3}$ \cite{adam}.

We trained each model on each regression dataset for 10,000 iterations, with the models fitting to a batch of 50 tasks during each iteration. Each batch consisted of new regression tasks sampled directly from our dataset's generator.

After training, we validated our model's posterior predictions using the NLL metric from \Cref{eq:metric} on sampled testing tasks, approximated with 20 Monte Carlo predictive posterior samples. We also measured an MSE metric by reducing the posterior samples of a model to a single prediction through averaging. While MSE reducing can be used to check our models are learning, it fails to measure the actual quality of predictive posteriors. We present MSE results in \Cref{sec:mse}.

Our models were implemented using the PyTorch and GPyTorch libraries \cite{torch,gpytorch}.

\subsection{Latent Gaussian Environment}
\label{sec:functionenv}
We propose a new, challenging function regression environment which directly models a function space which will have non-Gaussian predictive posteriors when conditioned on a few-shot training set. Each environment is parameterized by a fixed deterministic ``transform" function $f$, and a zero-mean Gaussian process $\mathcal G$ using a log-lengthscale $0.5$ RBF kernel, with a base log-variance sampled from $\mathcal N(0,1)$. To generate a $k$-shot task with $q$ query points, we sample $k+q$ values $x$ from $\mathcal N(0,1)$. We then sample $z\sim \mathcal G(x)$, and finally obtain $y=f(z)$ as the labels, so our generated task is $\D= (x,y)$.

We tested our method and baselines using each \[f\in \{\arctan(1/z), 5\operatorname{floor}(z), \tan(z), 5\sin(1/z)\}\] (clamping $|f|<10$ for the $\tan$ function). We chose these transform functions since they exhibit interesting global/non-continuous behavior, making it challenging for models to predict good posterior predictions. We used $k=10$ support points and $q=5$ query points for all these environments, except $\arctan(1/z)$, which used $k=5$ support points since it is smoother than the other tasks, and thus easier to learn well with a small $\D_\tr$. The results are shown in Table \ref{tab:latentnll}. We also report MSE results in Table \ref{tab:latentmse} in the Appendix.

\subsection{Standard Regression Environments}
\label{sec:sinenv}

\subsubsection{Sinusoid Regression}
\label{sec:ssinenv}

We also tested our model on two standard regression environments common in meta-learning literature \cite{maml, bmaml, dkt, mean}. In the ``Standard" sinusoid regression dataset, tasks were defined by sinusoid functions $y = A\sin(Bx+C)+\epsilon$, where our amplitude, frequency, and phase parameters were sampled uniformly from $A \in [0.1, 5.0]$, $B \in [0, 2\pi]$, and $C \in [0.5, 2.0]$. We then added observation noise $\epsilon \sim \N(0, (0.01A)^2)$ to each data point. A task was defined by sampling a value of $A, B,$ and $C$. The support and query datasets of a task were generated by sampling $k$ and $q$ $x$-values, respectively, uniformly randomly in $[-5.0, 5.0]$ and recording the input-label pairs $(x_i, A\sin(Bx_i+C) + \epsilon_i)$, where $\epsilon_i$ is noise sampled as described above.

We tested our model on three variants of this dataset.
\begin{itemize}
    \item  A ``High Frequency" sinusoid dataset, where $C$ was sampled uniformly from a larger range: $C \in [0.5, 15.0]$.
    \item  A ``Tangent" variant, which replaced the $\sin()$ with a $\tan()$ and clamped $y$ values between $-10.0$ and $10.0$.
    \item An ``Out of Range" variant, where $\M_{\rm tr}$ was generated normally, and test tasks $\M_{\rm test}$ were generated by sampling data points uniformly from $x \in [-5.0, 10.0]$.
\end{itemize}
We tested our models using $k=5$ support points on the ``Standard" dataset, and $k=10$ on the other datasets. We use a constant $q=5$ query points throughout. Negative log-likelihood results are shown in Table \ref{tab:sinnll}. We also report MSE in Table \ref{tab:sinmse} in the Appendix.


\subsubsection{Step Function Regression}
\label{sec:stepenv}
Lastly, we tested our model on step function regression, which is also common in meta-learning literature \cite{alpaca, mean}. In our step function environment, each task was defined by three points which were uniformly sampled from $[-2.5, 2.5]$. We labeled the samples $a, b$, and $c$ in increasing order, and our step function $y$ was defined for inputs $x$ as follows:
\begin{table}[H]
\begin{center}
    \begin{tabular}{l c c c c}
        \toprule
        & $x < a$ & $a \leq x < b$ & $b \leq x < c$ & $c \leq x$ \\
        \midrule
        $y$ & $-1 + \varepsilon$ & $1 + \varepsilon$ & $-1 + \varepsilon$ & $1 + \varepsilon$\\
        \bottomrule
    \end{tabular}
\end{center}
\end{table}
\vspace{-5mm}
where $\varepsilon \sim \N(0, 0.03^2)$ was noise sampled once per data point. Essentially, tasks in this dataset were functions that began at $-1$, switched abruptly between $-1$ and $1$ at three specified data points, and ended at $1$. The support and query datasets of a task were generated by sampling $k=5$ and $q=5$ $x$-values uniformly randomly in the range $[-5.0, 5.0]$, and recording the input-output pairs $(x_i, y_i)$.

We also tested on a ``high frequency" variant of this step function dataset where we had five switches instead of three in $[-2.5, 2.5]$, and our $y$ switched between $-2+\varepsilon$ and $2+\varepsilon$. We report NLL results on both step function variants in Table \ref{tab:stepnll}. We also report MSE in Table \ref{tab:stepmse} in the Appendix.
\begin{table}[htbp]
\begin{center}
    \begin{tabular}{l c c c c}
        \toprule
        & Standard & High Frequency \\
        \midrule
        EMAML & $3.671 \pm 0.037$ & $16.521 \pm 0.181$\\
        ALPaCA & $1.216 \pm 0.019$ & $2.861 \pm 0.071$\\
        DKT & $1.265 \pm 0.012$ & $2.072 \pm 0.022$\\
        VMGP (ours) & $\bm{0.453 \pm 0.013}$ & $\bm{0.742 \pm 0.063}$\\
        \bottomrule
    \end{tabular}
\end{center}
\caption{Average NLL $\pm$ Std. Error for Step Function Datasets.}
\label{tab:stepnll}
\end{table}

\section{Discussion}

\begin{figure*}[htpb]
  \centering
  \def\compwid{0.5}
  \def\fstskip{.48cm}
  \def\secskip{.67cm}
  \input{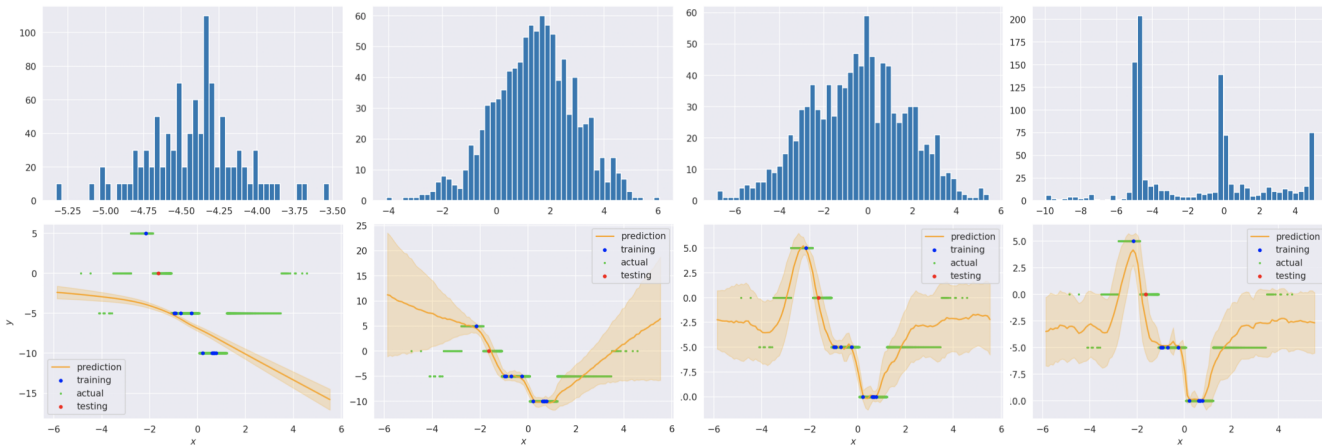}
  \vskip -2ex
  \caption{Predictions (bottom row) and posterior predictive distribution density (for the one red testing point, top row) from each model on a $5\operatorname{floor}(z)$ Latent Function task. The columns are, from left to right, EMAML (with 100 MAMLs), ALPaCA, DKT, and VMGP (ours).}
  \label{fig:predictposteriors}
\end{figure*}

Table \ref{tab:latentnll} shows our VMGP model outperforms all baselines on the four transform functions in the Latent Gaussian Environment. Looking qualitatively at the results, we see this performance arises directly from our model's ability to learn non-Gaussian posteriors. In Figure \ref{fig:predictposteriors}, we show an example of each model's predictions for a task from the $5\operatorname{floor}(z)$ dataset. The figures above each prediction show samples from the predictive posterior distribution conditioned on a testing point. We see that VMGP is the only method able to accurately represent this posterior as a multimodal mixture of point distributions: it learns that from the training data that the testing point is likely to either be $-5$, $0$, or $5$, rather than a sample from some Gaussian distribution like ALPaCA and DKT predict.

Table \ref{tab:sinnll} shows our model's NLL performance on datasets from the Sinusoid Regression Environment. We see that our model outperforms all others on the High Frequency and Tangent datasets, but ALPaCA bests it on Standard and Out of Range. Both DKT and ALPaCA also tend to outperform our model in terms of pure MSE accuracy (Table \ref{tab:sinmse}). We theorize these NLL results could be because the Standard/Out of Range sinusoids are relatively smooth functions, so the extra expressivity afforded by our model is simply not necessary for decent regression results. The High Frequency sinusoids and clamped tangent function are less Gaussian, giving our model an advantage. Example predictions by each model for Sinusoid Environment tasks are shown in Figure \ref{fig:regress} in the Appendix.

\begin{figure}[htpb]
  \centering
    \hspace*{-1.0cm}
  \begin{subfigure}[b]{0.5\columnwidth}
    \centering
    \includegraphics[scale=0.24]{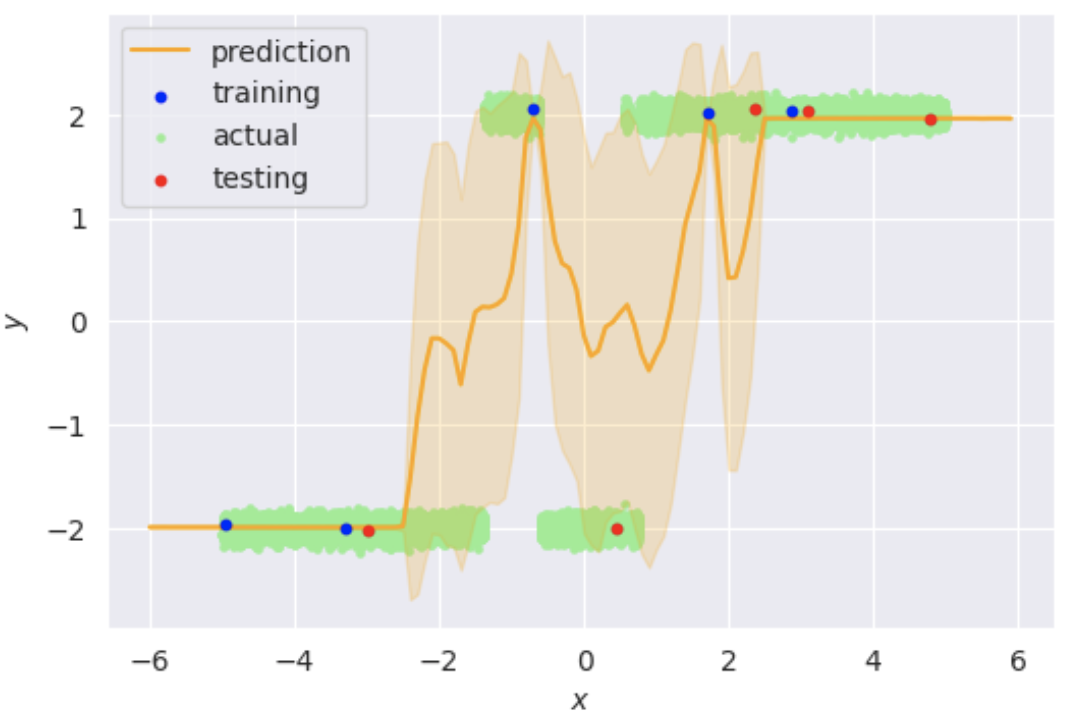}
  \end{subfigure}
  \hspace*{-0.4cm}
  \begin{subfigure}[b]{0.495\columnwidth}
    \centering
    \includegraphics[scale=0.345]{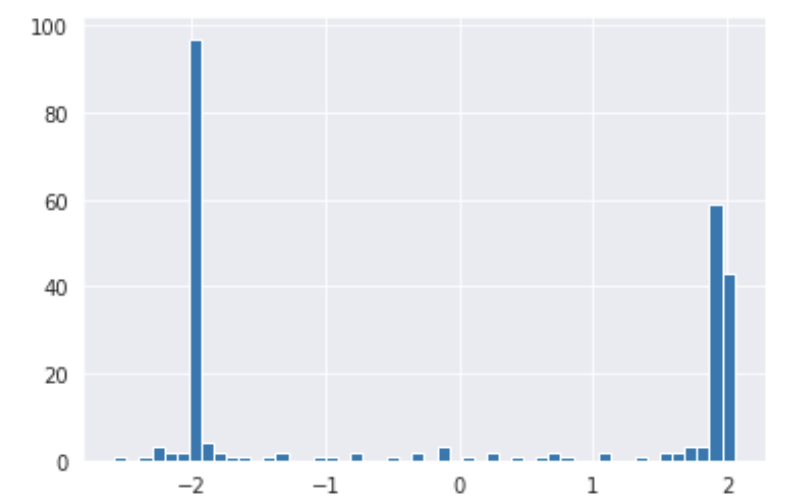}
  \end{subfigure}
  \caption{VGMP's predictions (left) and posterior predictive distribution density (for the red testing point at $x\approx0.5$) on a High Frequency Step Function Environment task.}
  \label{fig:stepposteriors}
\end{figure}

Table \ref{tab:stepnll} shows our model's NLL performance beats the baselines on both datasets from our Step Function Environment, and the margin of victory is larger on the more difficult high frequency dataset. Examining the predictive posterior (Figure \ref{fig:stepposteriors}) for a testing point, we again see a clear bimodal distribution, which no other method is able to express.

In general, VMGP's extra expressivity comes at the cost of slower learning, since our model has extra network parameters to learn. So, for simple sinusoidal regression datasets, running all models for the same number of iterations may naturally put ALPaCA and DKT ahead.

\section{Conclusion}

In this paper, we presented VMGP, a variational Gaussian-process-based meta-learning model. On most datasets, VMGP was able to predict significantly better posteriors for modeling its uncertainty compared to baseline methods. 

As measured by our NLL metric, VMGP significantly outperformed all other methods on our novel latent Gaussian environments described in \Cref{sec:functionenv}, as well as standard difficult trigonometric environments in \Cref{sec:sinenv} and the alternating step functions in \Cref{sec:ssinenv}.

Qualitative examinations of the posterior predictions confirmed that when regressing functions with discontinuities or less smooth behavior, VMGP was able to produce visibly multimodal and skewed non-Gaussian posterior predictive distributions.

However, on simple sinusoid regression tasks, the ALPaCA algorithm did outperform VMGP. These smooth functions are well-modeled with Gaussian posteriors, so the extra expressivity of our model was not needed and our longer training time hurt performance.

\subsection{Future Work}
Additional architectural tuning for the hyperparameters of the VMGP method could likely improve our results. Specifically, given the complexity of our VMGP architecture, which jointly trains three models $\p,\f,\q$, it may be beneficial to experiment with lower learning rates and higher numbers of training iterations than the defaults of $10^{-3}$ and 10,000 respectively. We also used the same 2-layer 40-hidden-unit MLP architecture for $\p,\f,\q$, so it may be beneficial to experiment with varying the individual (or overall) deep model architectures.

Additional improvements to VMGP may come from combining it with more complex kernels than the current RBF kernel. Patacchiola et al. \yrcite{dkt} found that their deep kernel transfer meta-learning was more effective with a spectral kernel in certain environments, and similar benefits would likely be conferred on VMGP. Furthermore, by combining VMGP with a linear kernel and mean function, VMGP could learn nonlinear transforms of Bayesian linear regression, yielding a combined VMGP+ALPaCA method that may perform better in environments where ALPaCA does well. For some difficult periodic functions for which all methods struggle to obtain accurate predictions (cf. Figure \ref{fig:regress}), testing combining our methods with kernels that confer knowledge of a periodic prior could yield improvements.

Robustness of VMGP could also be tested by training on a real-world image-based pose prediction task \cite{pose}. By replacing the MLP models in VMGP with CNNs, we can run the VMGP algorithm directly on these tasks.


\bibliography{sources}
\bibliographystyle{icml2020}

\appendix
\newpage
\onecolumn

\section{Code}

Our code is publicly available at \url{https://github.com/vivekmyers/vmgp}.
An interactive notebook illustrating our methods can be accessed at \url{https://colab.research.google.com/drive/1TGWta5PcgEy0C6oBQOGHsNlpGi4Hnup6?usp=sharing}.
\section{MSE Results}
\label{sec:mse}
Tables \ref{tab:latentmse}, \ref{tab:sinmse}, and \ref{tab:stepmse} show our MSE results on the Latent Gaussian Environment, Sinusoid Environment, and Step Function Environment datasets.
\begin{table}[H]
\begin{center}
    \begin{tabular}{l c c c c}
        \toprule
        & $\arctan(1/z)$ & floor$(z)$ & $\tan(z)$ & $\sin(1/z)$ \\
        \midrule
        EMAML & $0.586 \pm 0.011$ & $8.496 \pm 0.153$ & $9.684 \pm 0.201$ & $8.793 \pm 0.093$ \\
        ALPaCA & $0.503 \pm 0.013$ & $2.783 \pm 0.052$ & $9.077 \pm 0.217$ & $\bm{6.273\pm 0.089}$\\
        DKT & $\bm{0.488 \pm 0.0129}$ & $\bm{2.694 \pm 0.051}$ & $8.212 \pm 0.189$ & $6.355 \pm 0.079$ \\
         VMGP (ours) & $0.581 \pm 0.022$ & $3.487 \pm 0.102$ & $\bm{7.922 \pm 0.288}$ & $13.556 \pm 0.116$ \\
        \bottomrule
    \end{tabular}
\end{center}
\caption{Average MSE $\pm$ Std. Error for Latent Gaussian Environment Datasets.}
\label{tab:latentmse}
\end{table}

\begin{table}[H]
\begin{center}
    \begin{tabular}{l c c c c}
        \toprule
        & Standard & High Frequency & Out of Range & Tangent \\
        \midrule
        EMAML & $4.145 \pm 0.097$ & $4.268 \pm 0.092$ & $12.619 \pm 0.693$ & $25.648 \pm 0.462$ \\
        ALPaCA & $\bm{1.209 \pm 0.044}$ & $4.165 \pm 0.091$ & $\bm{0.198 \pm 0.020}$ & $26.761 \pm 0.582$\\
        DKT & $2.021 \pm 0.069$ & $\bm{3.745 \pm 0.084}$ & $1.556 \pm 0.065$ & $\bm{22.204 \pm 0.440}$ \\
        VMGP (ours) & $2.210 \pm 0.075$ & $3.764 \pm 0.172$ & $1.882 \pm 0.170$ & $24.350 \pm 0.861$ \\
        \bottomrule
    \end{tabular}
\end{center}
\caption{Average MSE $\pm$ Std. Error for Sinusoid Environment Datasets.}
\label{tab:sinmse}
\end{table}

\begin{table}[htbp]
\begin{center}
    \begin{tabular}{l c c c c}
        \toprule
        & Standard & High Frequency \\
        \midrule
        EMAML & $0.395 \pm 0.004$ & $2.110 \pm 0.023$\\
        ALPaCA & $\bm{0.377 
        \pm 0.006}$ & $\bm{1.841 \pm 0.027}$\\
        DKT & $0.527 \pm 0.005$ & $2.214 \pm 0.018$\\
        VMGP (ours) & $0.474 \pm 0.007$ & $2.059 \pm 0.029$\\
        \bottomrule
    \end{tabular}
\end{center}
\caption{Average MSE $\pm$ Std. Error for Step Function Environment Datasets.}
\label{tab:stepmse}
\end{table}

\onecolumn
\newcommand\wid{0.24}
\newcommand\scalef{0.35}
\section{Regression Examples}
\begin{figure*}[htbp]
\everypar{\hskip -2.5em}
  \centering
  \begin{subfigure}[b]{\wid\columnwidth}
    \centering
    \includegraphics[scale=\scalef]{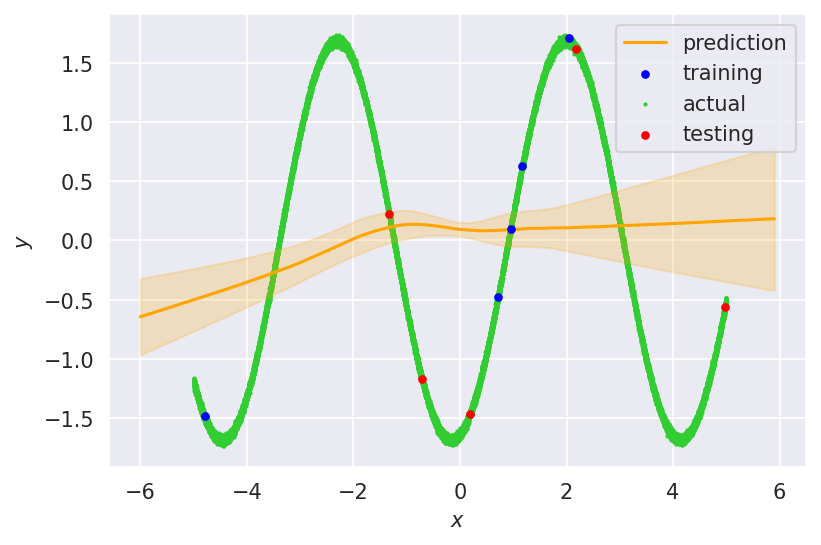}
  \end{subfigure}
  \hfill
  \begin{subfigure}[b]{\wid\columnwidth}
    \centering
    \includegraphics[scale=\scalef]{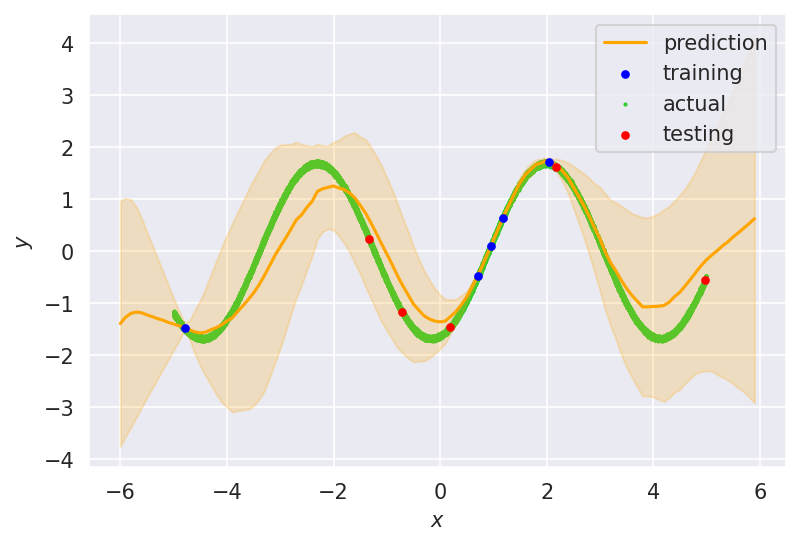}
  \end{subfigure}
  \hfill 
  \begin{subfigure}[b]{\wid\columnwidth}
    \centering
    \includegraphics[scale=\scalef]{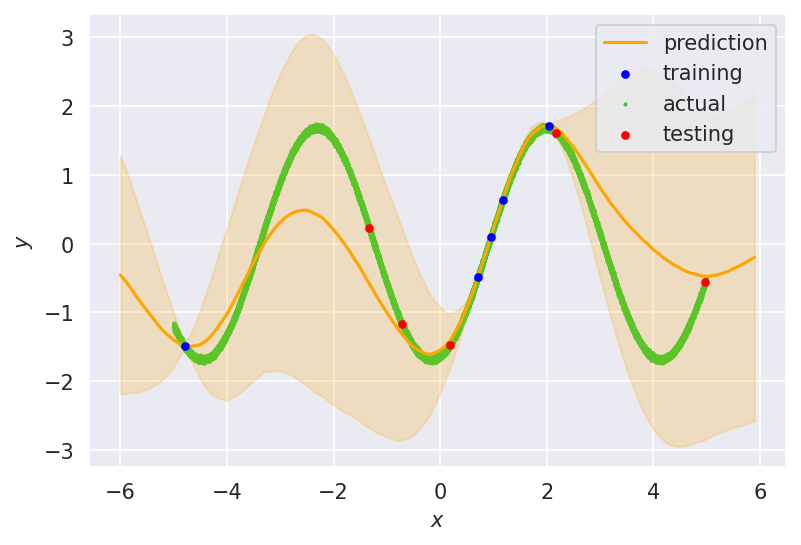}
  \end{subfigure}
  \hfill 
    \begin{subfigure}[b]{\wid\columnwidth}
    \centering
    \includegraphics[scale=\scalef]{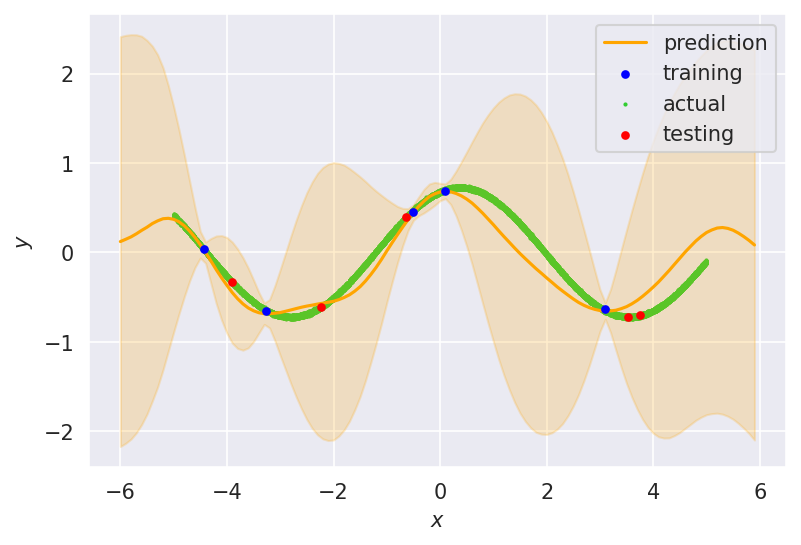}
  \end{subfigure}
  
  \begin{subfigure}[b]{\wid\columnwidth}
    \centering
    \includegraphics[scale=\scalef]{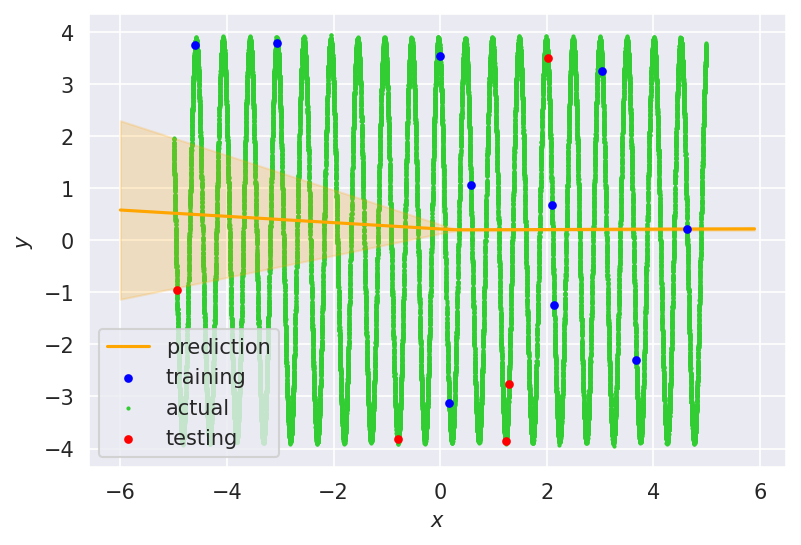}
  \end{subfigure}
  \hfill
  \begin{subfigure}[b]{\wid\columnwidth}
    \centering
    \includegraphics[scale=\scalef]{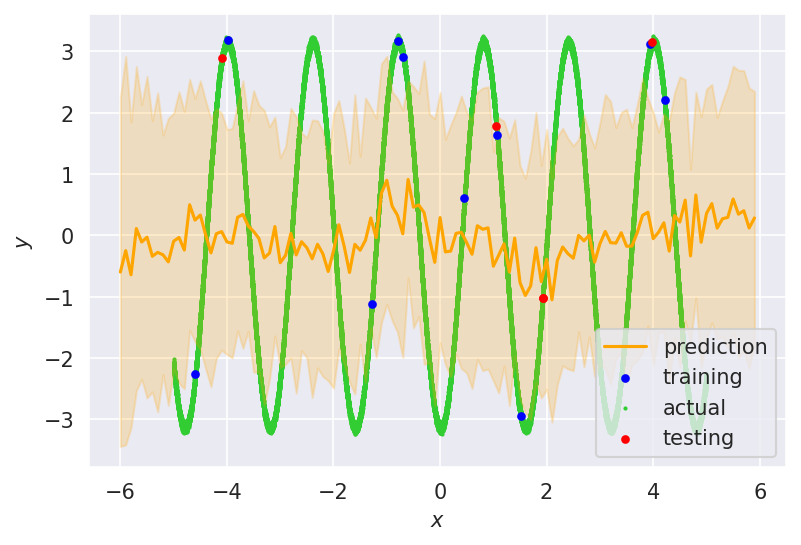}
  \end{subfigure}
  \hfill 
  \begin{subfigure}[b]{\wid\columnwidth}
    \centering
    \includegraphics[scale=\scalef]{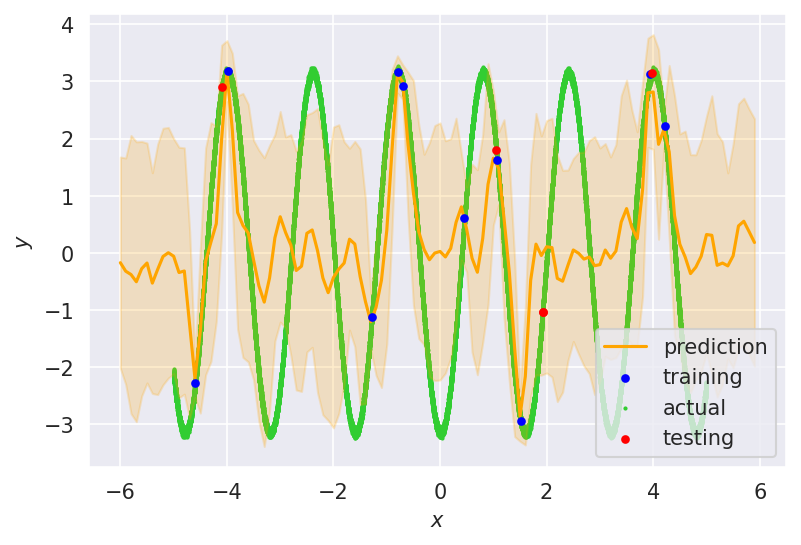}
  \end{subfigure}
  \hfill 
    \begin{subfigure}[b]{\wid\columnwidth}
    \centering
    \includegraphics[scale=\scalef]{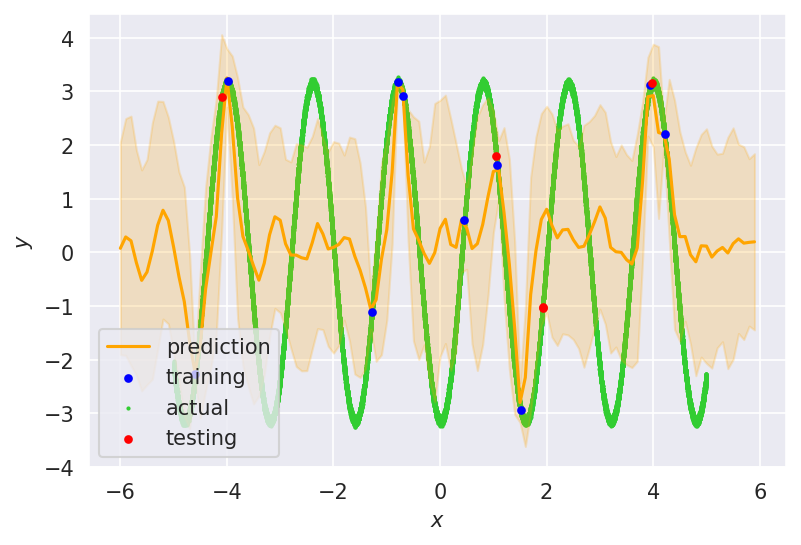}
  \end{subfigure}

  \begin{subfigure}[b]{\wid\columnwidth}
    \centering
    \includegraphics[scale=\scalef]{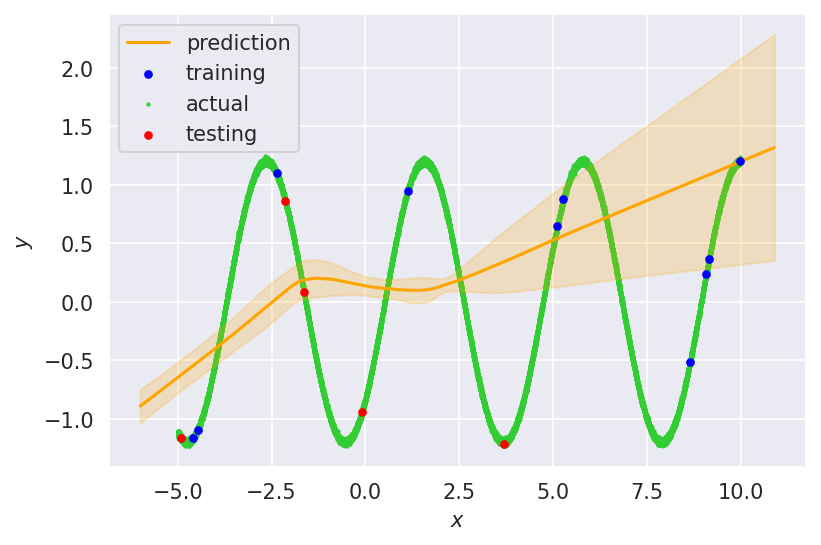}
  \end{subfigure}
  \hfill
  \begin{subfigure}[b]{\wid\columnwidth}
    \centering
    \includegraphics[scale=\scalef]{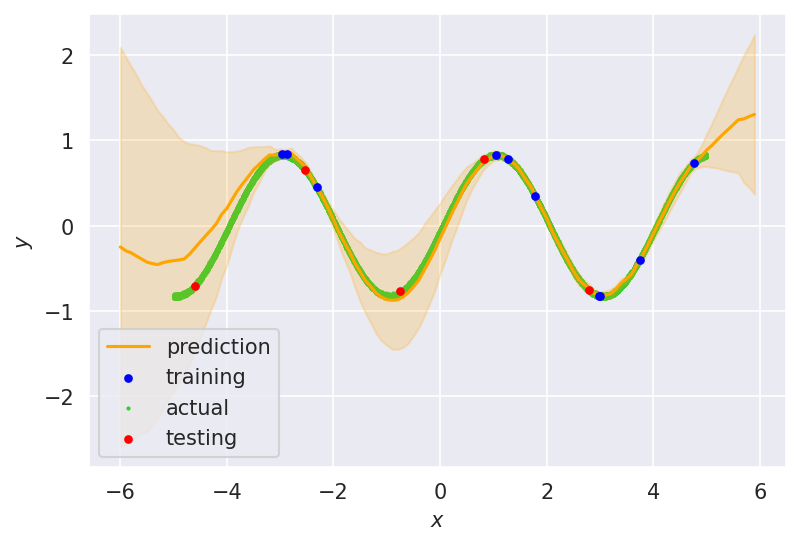}
  \end{subfigure}
  \hfill 
  \begin{subfigure}[b]{\wid\columnwidth}
    \centering
    \includegraphics[scale=\scalef]{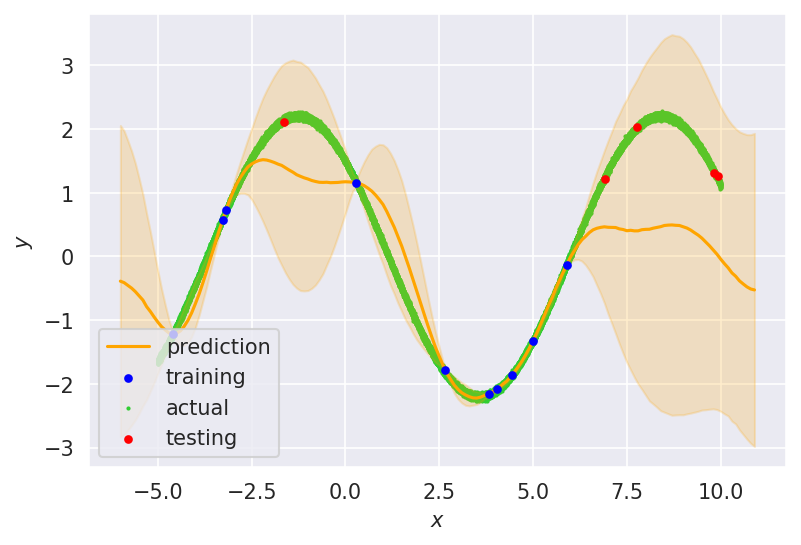}
  \end{subfigure}
  \hfill 
    \begin{subfigure}[b]{\wid\columnwidth}
    \centering
    \includegraphics[scale=\scalef]{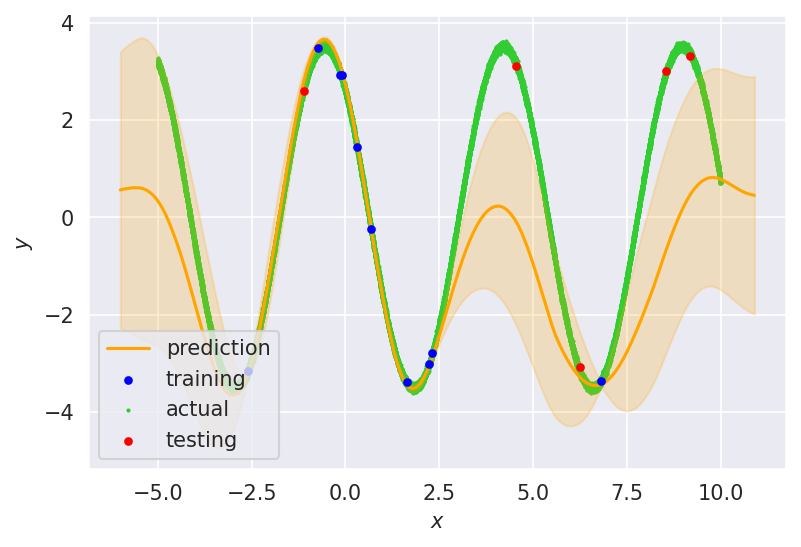}
  \end{subfigure}
  
\begin{subfigure}[b]{\wid\columnwidth}
    \centering
    \includegraphics[scale=\scalef]{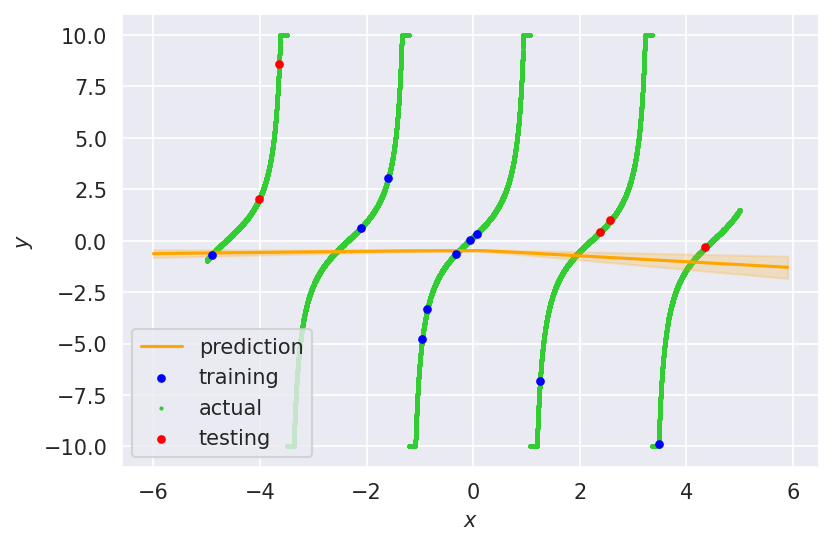}
  \end{subfigure}
  \hfill
  \begin{subfigure}[b]{\wid\columnwidth}
    \centering
    \includegraphics[scale=\scalef]{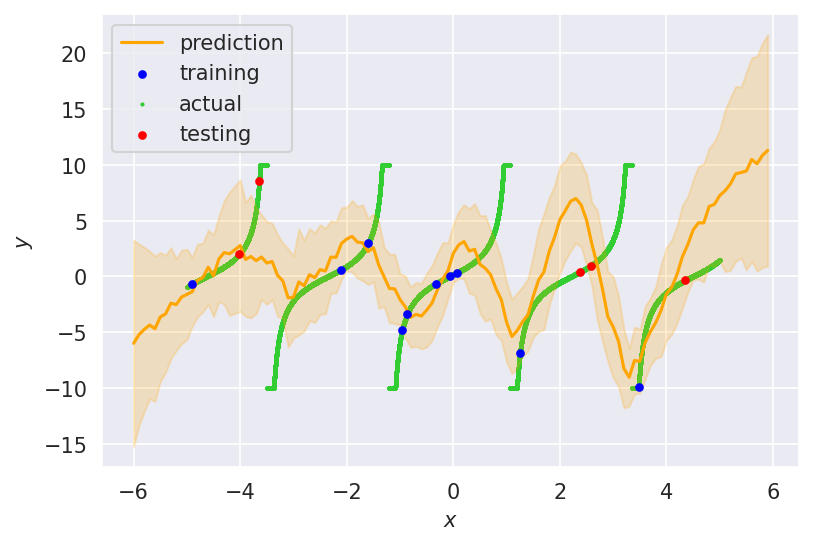}
  \end{subfigure}
  \hfill 
  \begin{subfigure}[b]{\wid\columnwidth}
    \centering
    \includegraphics[scale=\scalef]{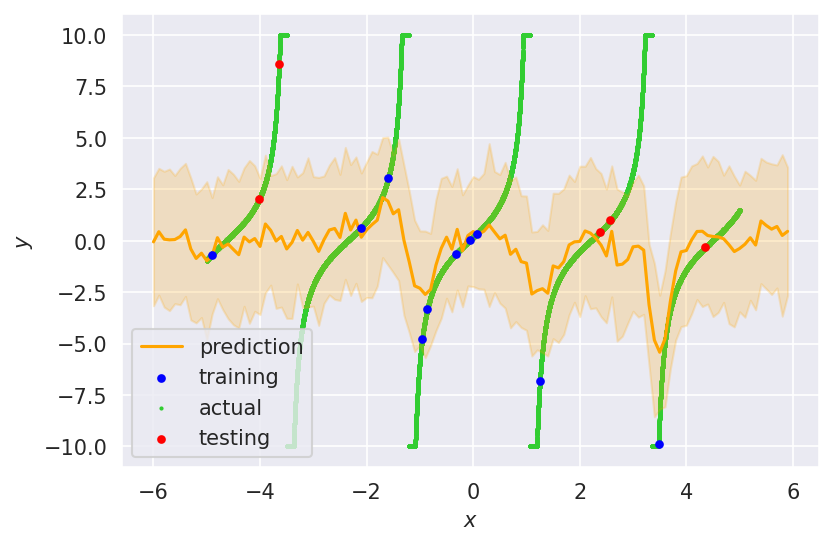}
  \end{subfigure}
  \hfill 
    \begin{subfigure}[b]{\wid\columnwidth}
    \centering
    \includegraphics[scale=\scalef]{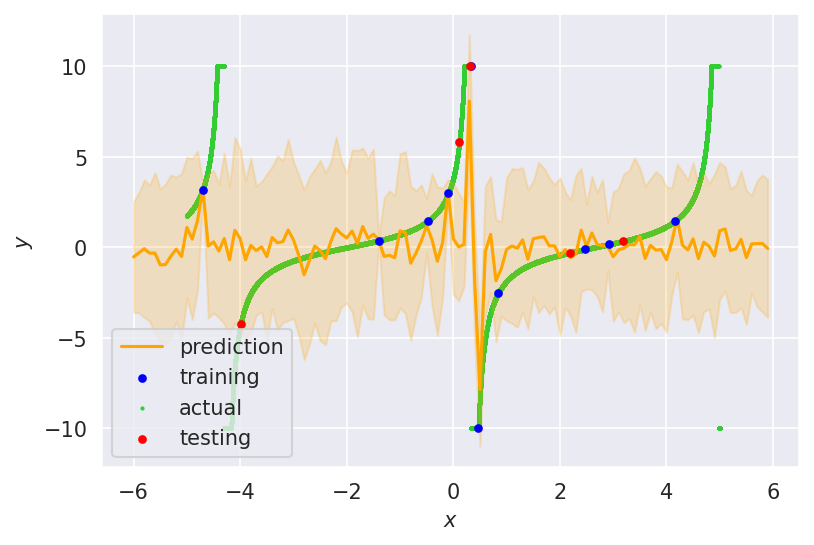}
 \end{subfigure}
  \caption{Sinusoid Regression Examples. Columns, from left: EMAML, AlPaCA, DKT, VMGP (ours). Each row shows an example test task from a different variant of the sinusoid environment. Rows, from top: Standard, High Frequency, Out of Range, Tangent.}
\label{fig:regress}
\end{figure*}

\end{document}